\documentclass[a4paper]{jpconf}
\usepackage{graphicx}
\newtheorem{thm}{Theorem}
\usepackage{bm}
\begin{document}
\title{Quantum field theories, Markov random fields and machine learning}

\author{Dimitrios Bachtis$^1$, Gert Aarts$^{2,3}$ and Biagio Lucini$^{1,4}$}
\address{$^1$Department of Mathematics,  Swansea University, Bay Campus, SA1 8EN, Swansea, Wales, United Kingdom}
\address{$^2$Department of Physics, Swansea University, Singleton Campus, SA2 8PP, Swansea, Wales, United Kingdom}
\address{$^3$European Centre for Theoretical Studies in Nuclear Physics and Related Areas (ECT*) \& Fondazione Bruno Kessler
Strada delle Tabarelle 286, 38123 Villazzano (TN), Italy }
\address{$^4$Swansea Academy of Advanced Computing, Swansea University, Bay Campus, SA1 8EN, Swansea, Wales, United Kingdom}

\ead{dimitrios.bachtis@swansea.ac.uk, g.aarts@swansea.ac.uk, b.lucini@swansea.ac.uk}

\begin{abstract}
The transition to Euclidean space and the discretization of quantum field theories on spatial or space-time lattices opens up the opportunity to investigate probabilistic machine learning within quantum field theory. Here, we will discuss how discretized Euclidean field theories, such as the $\phi^{4}$ lattice field theory on a square lattice, are mathematically equivalent to Markov fields, a notable class of probabilistic graphical models with applications in a variety of research areas, including machine learning. The results are established based on the Hammersley-Clifford theorem. We will then derive neural networks from quantum field theories and discuss applications pertinent to the minimization of the Kullback-Leibler divergence for the probability distribution of the $\phi^{4}$ machine learning algorithms and other probability distributions.
\end{abstract}

\section{Introduction}

To construct a probability distribution in a high-dimensional space one can turn to the framework of probabilistic graphical models. Probabilistic graphical models comprise a set of random variables, positioned within a graph-based representation, that satisfy certain factorization as well as conditional dependence and independence properties. One notable case of probabilistic graphical models is a Markov field, in which the random variables are connected through undirected edges and satisfy a significant condition of locality, called the Markov property. Markov properties emerge as important mathematical conditions across distinct research fields, such as in machine learning~\cite{koller-pgm} or in constructive quantum field theory~\cite{nelson-qft}.

In this contribution, we discuss the proof of Markov properties for discretized Euclidean field theories \cite{bachtis-qftml}. Specifically, we demonstrate, through the Hammersley-Clifford theorem, that the $\phi^{4}$ scalar field theory on a square lattice satisfies the local Markov property, and is therefore mathematically equivalent to a Markov field. Based on this equivalence, we introduce algorithms which generalize a notable class of neural networks, specifically restricted Boltzmann machines. Finally, we present applications pertinent to the minimization of the Kullback-Leibler divergence for the probability distribution of the $\phi^{4}$ machine learning algorithms and other probability distributions.

\section{The $\phi^{4}$ Markov field}

\begin{figure}[b]
\center
\includegraphics[width=6.5cm]{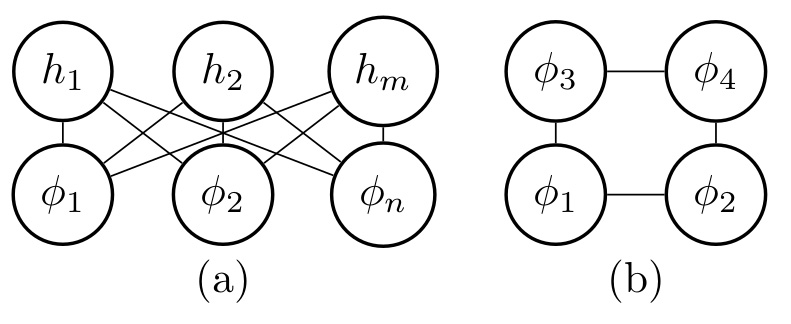}
\caption{\label{fig:graph} A bipartite graph (a) and a square lattice (b). Examples of maximal cliques are $\lbrace \phi_{1}, h_{1} \rbrace$ and $\lbrace \phi_{3}, \phi_{4} \rbrace$, respectively. }
\end{figure}

We denote as $\Lambda$ a finite set which is equivalently expressed as a graph $\mathcal{G}=(\Lambda,e)$, where the points of $\Lambda$ correspond to the vertices of $\mathcal{G}$ and $e$ denotes the edges of the graph. Two vertices $i,j \in \Lambda$ which are connected by an edge are neighbours. A clique is a set of neighbours, which is called maximal if no additional vertex can be included that is simultaneously a neighbour with all the vertices present in the clique, see Fig.~\ref{fig:graph}. We now assign to each vertex $i$ in the graph $\mathcal{G}$ a continuous-valued random variable, which is denoted as $\phi_{i}$.

A Markov random field is defined as a set of random variables on a graph $\mathcal{G}=(\Lambda,e)$ whose associated probability distribution $p(\phi)$ satisfies the local Markov property:
\begin{equation}
p(\phi_{i} | (\phi_{j})_{j \in \Lambda - i}) = p(\phi_{i} | (\phi_{j})_{j \in \mathcal{N}_{i}}),
\end{equation}
where $\mathcal{N}_{i}$ is the set of neighbours of a given point $i$. The local Markov property can be proven for a probability distribution that is encoded in a graph through the Hammersley-Clifford theorem:

\begin{thm}[Hammersley-Clifford]
A probability distribution $p$, satisfying the condition of positivity, is associated with the events generated by a Markov network, iff $p$ can be factorized as a product of positive factors, or potential functions $\psi_{c}$, over the cliques of the associated graph structure $\mathcal{G}$:
\begin{equation}
p(\phi)= \frac{1}{Z} \prod_{c \in C} \psi_{c}(\phi),
\end{equation}
where $Z=\int_{\bm{\phi}} \prod_{c \in C} \psi_{c}(\bm{\phi}) d\bm{\phi}$ is a normalization constant, $c \in C$ is a maximal clique,  and $\bm{\phi}$ denotes all configurations of the system.

\end{thm}

The Euclidean action of the two-dimensional $\phi^{4}$ scalar field theory is
\begin{equation}\label{eq:midaction}
S_{E}= -\kappa_{L}\sum_{\langle ij \rangle}\phi_{i} \phi_{j} + \frac{(\mu_{L}^{2}+4\kappa_{L})}{2} \sum_{i} \phi_{i}^{2} +  \frac{\lambda_{L}}{4}  \sum_{i}\phi_{i}^{4},
\end{equation}
where  $\kappa_{L},\mu_{L}^{2},\lambda_{L}$ are dimensionless parameters. We redefine $w=\kappa_{L}$, $a=(\mu_{L}^{2}+4\kappa_{L})/2$, $b=\lambda_{L}/4$ for simplicity, and consider them as inhomogeneous:
\begin{equation}\label{eq:finalaction}
S(\phi ; \theta)= -\sum_{\langle ij \rangle} w_{ij} \phi_{i}\phi_{j} + \sum_{i} a_{i} \phi_{i}^{2} + \sum_{i} b_{i} \phi_{i}^{4}.
\end{equation}

The $\phi^{4}$ inhomogeneous action is described by the coupling constants $\theta=\lbrace w_{ij}, a_{i},b_{i} \rbrace$, and the Boltzmann probability distribution:
\begin{equation} \label{eq:probmrf}
p(\phi ; \theta)=\frac{\exp\big[-S(\phi ; \theta)\big]}{\int_{\bm{\phi}}{\exp[-S(\bm{\phi},\theta)]} d\bm{\phi}}.
\end{equation}
The lattice version of the $\phi^{4}$ theory is, by definition, expressed as a graph. To verify that the $\phi^{4}$ theory satisfies Markov properties we define the following potential function $\psi_{c}$ that is able to factorize the probability distribution in terms of maximal cliques $c \in C$
\begin{equation}
\psi_{c} = \exp\bigg[ -w_{ij} \phi_{i} \phi_{j}+ \frac{1}{4} (a_{i} \phi_{i}^{2} +a_{j}\phi_{j}^{2}  +b_{i} \phi_{i}^{4} +b_{j}\phi_{j}^{4})\bigg],
\end{equation}
where $i,j$ are nearest neighbours.

\section{Machine learning with $\phi^{4}$ Markov random fields}

\subsection{Learning without predefined data}

To compare the probability distribution $p(\phi; \theta)$ of the Markov field with another probability distribution $q(\phi)$, we define the Kullback-Leibler divergence, a nonnegative quantity, as
\begin{equation} \label{eq:kl}
KL(p || q) = \int_{-\infty}^{\infty} {p(\bm{\phi}; \theta)} \ln \frac{p(\bm{\phi}; \theta)}{q(\bm{\phi})} d\bm{\phi}  \geq 0.
\end{equation}

We now consider a target Boltzmann probability distribution $q(\phi)= \exp [-\mathcal{A}]/Z_{\mathcal{A}}$ that describes an arbitrary statistical system and substitute the two probability distributions in the Kullback-Leibler divergence to obtain:
\begin{equation}\label{eq:fen2}
F_{\mathcal{A}} \leq \langle \mathcal{A} - S \rangle_{p(\phi;\theta)} + F \equiv \mathcal{F},
\end{equation}
where $\mathcal{F}$ is the variational free energy,  $F_{\mathcal{A}}=-\ln Z_{\mathcal{A}}$, and $\langle O \rangle_{p(\phi;\theta)}$ denotes the expectation value of an observable $O$ under the probability distribution $p(\phi;\theta)$. By minimizing this quantity the two probability distributions  $p(\phi;\theta)$  and $q(\phi)$ will become equal and we can therefore use the distribution of the $\phi^{4}$ theory to draw samples from the target probability distribution $q(\phi)$.

Based on a gradient-based approach, we are able to minimize Eq.~\ref{eq:fen2} via
\begin{equation}
\frac{\partial \mathcal{F}}{\partial \theta_{i}}= \langle \mathcal{A} \rangle \Big\langle \frac{\partial S}{\partial \theta_{i}} \Big\rangle -\Big\langle \mathcal{A} \frac{\partial S}{\partial \theta_{i}} \Big\rangle + \Big\langle S \frac{\partial S}{\partial \theta_{i}} \Big\rangle - \langle S \rangle \Big\langle \frac{\partial S}{\partial \theta_{i}} \Big\rangle,
\end{equation} 
and the coupling constants $\theta$ are updated at each epoch $t$ as:
\begin{equation}\label{eq:gas}
\theta^{(t+1)}=\theta^{(t)}-\eta*  \mathcal{L},
\end{equation}
where $\eta$ is the learning rate and $\mathcal{L}=\partial \mathcal{F}/\partial \theta^{(t)}$.

We now consider a variation of the $\phi^{4}$ theory which includes next-nearest neighbor interactions $nnn$,  and with a complex action $\mathcal{A}$ defined as 
\begin{equation} \label{eq:fullaction}
\mathcal{A}= \sum_{k=1}^{5} g_{k}\mathcal{A}^{(k)}=  g_{1} \sum_{\langle ij \rangle_{nn} } \phi_{i} \phi_{j} + g_{2} \sum_{i} \phi_{i}^{2}  +  g_{3} \sum_{i} \phi_{i}^{4}  + g_{4} \sum_{\langle ij \rangle_{nnn} } \phi_{i} \phi_{j} + i g_{5} \sum_{i}  \phi_{i}^{2},
\end{equation}
where $i$ denotes the imaginary unit. The coupling constants can have arbitrary values but for this example we consider $g_{1}=g_{4}=-1$, $g_{2}=1.52425$, $g_{3}=0.175$ and $g_{5}=0.15$, see Ref.~\cite{bachtis-qftml}. 

\begin{figure}[t]
\includegraphics[width=7.5cm]{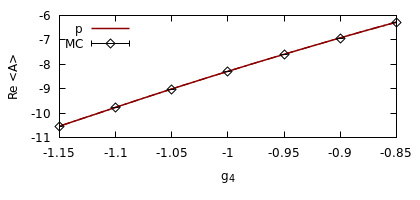}
\includegraphics[width=7.5cm]{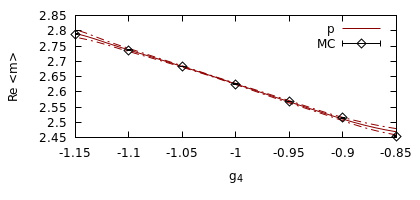}
\caption{\label{fig:re1} $\Re [\mathcal{A}]$ (left) and $\Re [m]$ (right) versus $g_{4}$.  The statistical errors are comparable with the width of the line. }
\end{figure}

We now utilize the $\phi^{4}$ Markov field of action $S$, to approximate a probability distribution which is described by an action $\mathcal{A}_{\lbrace 4\rbrace }=\sum_{k=1}^{4} g_{k}\mathcal{A}^{(k)}$. To investigate how accurate the equivalence between the two probability distributions is, we will implement reweighting based on the probability distribution of the $\phi^{4}$ Markov field to calculate expectation values of the full complex action $\mathcal{A}$. The reweighting relation is given by
\begin{equation}\label{eq:rewfull}
\langle O \rangle= \frac{\sum_{l=1}^{N} O_{{l}} \exp[S_{{l}}-g_{j}'\mathcal{A}_{{l}}^{(j)}- \sum_{k=1,k \neq j}^{5}g_{k}\mathcal{A}^{(k)}_{{l}}]}{\sum_{l=1}^{N}  \exp[S_{{l}}-g_{j}'\mathcal{A}_{{l}}^{(j)}- \sum_{k=1,k \neq j}^{5}g_{k}\mathcal{A}_{{l}}^{(k)}]},
\end{equation}
details of which can be found in Refs.~\cite{bachtis-qftml,bachtis-addml,bachtis-map,bachtis-ext}.

We consider $j=4$ and calculate the expectation values of the action $\mathcal{A}$ and the magnetization $m$  by extrapolating with reweighting in the range $g_{4}' \in[-1.15,-0.85]$. The results, shown in Fig.~\ref{fig:re1}, overlap within statistical uncertainty with independent calculations, therefore verifying that observables which would correspond to probability distribution of the full action $\mathcal{A}$ can be accurately calculated based on reweighting from the inhomogeneous action $S$.

\subsection{Learning with predefined data}

A different class of machine learning applications considers the case where the form of the target probability distribution is unknown but there exists a set of available data in which an empirical probability distribution is encoded. To explore this type of applications we consider the following expression of the Kullback-Leibler divergence:

\begin{equation} \label{eq:klopp}
KL(q || p) = \int_{-\infty}^{\infty} {q(\bm{\phi})} \ln \frac{q(\bm{\phi})}{p(\bm{\phi}; \theta)} d\bm{\phi}  \geq 0.
\end{equation}

By substituting and taking the derivative in terms of the variational parameters $\theta$ we obtain:
\begin{equation}
\frac{\partial \ln p(\phi ; \theta)}{\partial \theta} = \Big\langle \frac{\partial S}{\partial \theta} \Big\rangle_{p(\phi ; \theta)} -  \frac{\partial S}{\partial \theta},
\end{equation}
and the update rule of the parameters $\theta$  at each epoch $t$ is given based on Eq.~\ref{eq:gas}, where $\mathcal{L}=-\partial \ln p(\phi ; \theta^{(t)})/{\partial \theta^{(t)}}$.

\begin{figure}[t]
\includegraphics[width=14.2cm]{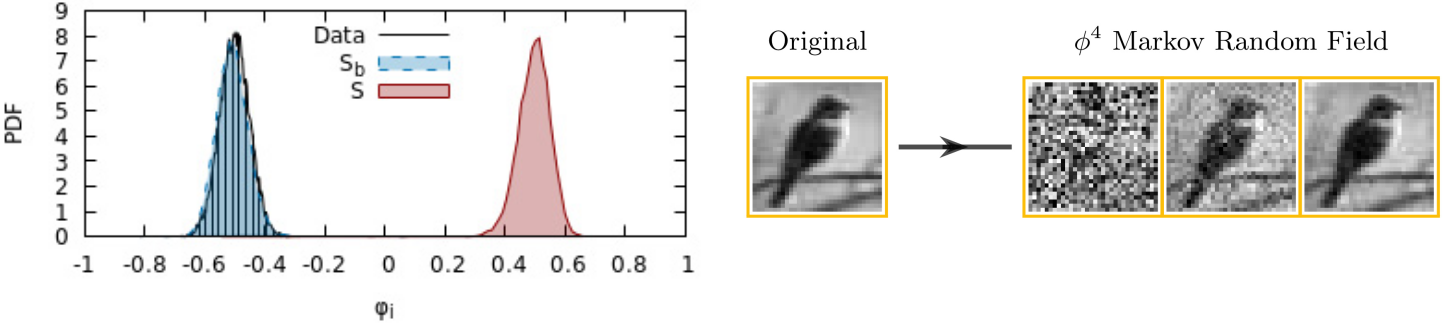}
\caption{\label{fig:2} Probability density function (PDF) versus $\phi_{i}$ (left). The thermalization of the trained Markov field (right).}
\end{figure}

As an example of a distribution to be learned by the machine learning algorithm, we consider the simple case of a Gaussian distribution with $\mu=-0.5$ and $\sigma=0.05$. Since the lattice action is invariant under the $Z_{2}$ symmetry we expect that the symmetric values of the dataset are equiprobable in being reproduced. This invariance can be removed via the inclusion of a term $\sum_{i} r_{i} \phi_{i}$ which breaks the symmetry of the action explicitly. The results are shown in Fig.~\ref{fig:2} (left) where the anticipated behaviour is observed.

Finally, we illustrate the approach in an image from the CIFAR-10 dataset. The thermalization of the trained Markov field is depicted in Fig.~\ref{fig:2} (right), where the image emerges within the equilibrium probability distribution. Since the machine learning algorithm learns the correct values of coupling constants in the action that solve the considered problem, extensions towards learning the appropriate coupling constants which describe renormalized systems can potentially be explored~\cite{bachtis-irg}.

\subsection{Machine learning with $\phi^{4}$ neural networks}

To increase the expressivity of the machine learning algorithm, we will introduce a new set of latent or hidden variables $h_{j}$ within the graph structure. In addition, we will restrict the interactions to be exclusively between the visible $\phi_{i}$ and the hidden $h_{j}$ variables, giving rise to the lattice action  

\begin{figure}[b]
\center
\includegraphics[width=8.6cm]{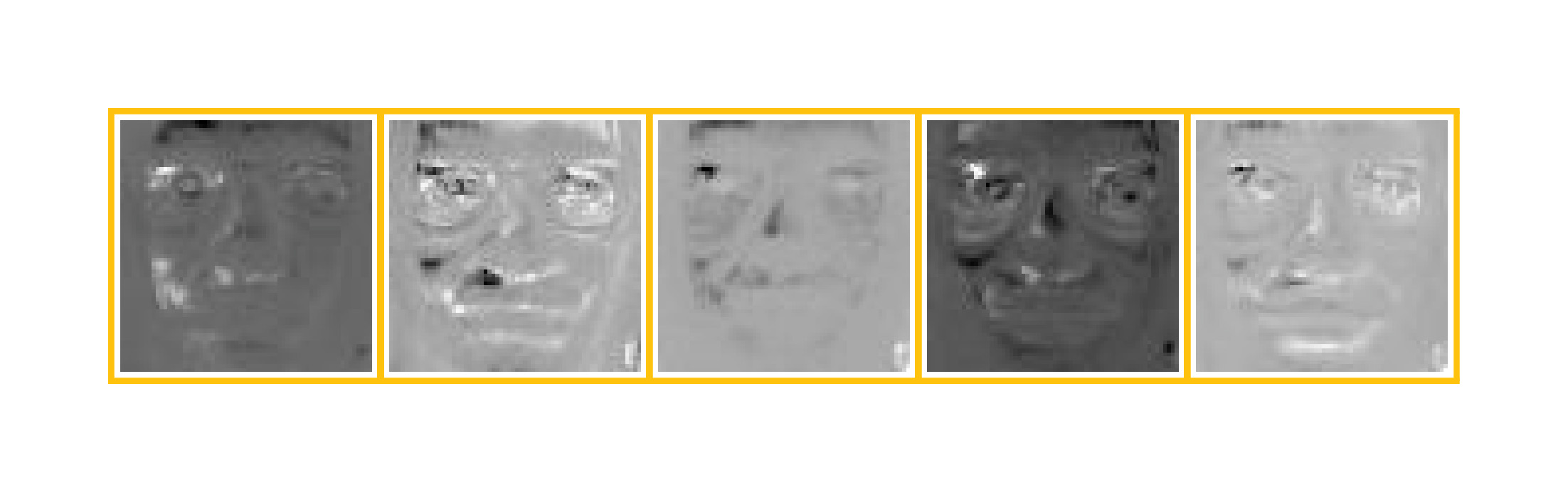}
\caption{\label{fig:faces} Extracted dependencies from the trained $\phi^{4}$ neural network.}
\end{figure}

\begin{equation}\label{eq:many}
S (\phi,h ; \theta) =  -\sum_{i,j} w_{ij} \phi_{i}h_{j}  + \sum_{i} r_{i} \phi_{i}  + \sum_{i} a_{i} \phi_{i}^{2} + \sum_{i} b_{i} \phi_{i}^{4}  + \sum_{j} s_{j} h_{j} + \sum_{j} m_{j} h_{j}^{2} + \sum_{j} n_{j} h_{j}^{4}.
\end{equation}

The $\phi^{4}$ neural network implemented above is a generalization of other neural network architectures: if $b_i=n_j=0$ the $\phi^{4}$ neural network reduces to a Gaussian-Gaussian restricted Boltzmann machine and if $b_i=n_j=m_j=0$ and $h_{j} \in \lbrace-1,1\rbrace$ then one obtains a Gaussian-Bernoulli restricted Boltzmann machine, see Refs~\cite{hinton-pgm,fischer-pgm}. Restricted Boltzmann machines were inspired by Ising models and since the $\phi^{4}$ theory can become equivalent to an Ising model~\cite{milchev-phi}, novel physical insights into how these machine learning algorithms work might emerge from the perspective of quantum field theory. We now train the  $\phi^{4}$ neural network on the Olivetti faces dataset. Some representative features, which resemble face structures, are shown in Fig.~\ref{fig:faces}.

\section{Conclusions}

In this contribution we have shown that discretized Euclidean field theories are Markov fields, hence it is now possible to investigate machine learning directly within quantum field theory~\cite{bachtis-qftml}. Based on the Hammersley-Clifford theorem, we demonstrated that the $\phi^{4}$ theory, formulated on a square lattice, satisfies the local Markov property. In addition we have derived $\phi^{4}$ neural networks which generalize restricted Boltzmann machines (see also Refs.~\cite{hash-adscft,hash-adscft2}). Finally, we have presented applications pertinent to physics and computer science, opening up the opportunity for nonperturbative investigations of machine learning within lattice field theory.

\ack
The authors received funding from the European Research Council (ERC) under the European Union's Horizon 2020 research and innovation programme under grant agreement No 813942. The work of GA and BL has been supported in part by the UKRI Science and Technology Facilities Council (STFC) Consolidated Grant ST/T000813/1. The work of BL is further supported in part by the Royal Society Wolfson Research Merit Award WM170010 and by the Leverhulme Foundation Research Fellowship RF-2020-461\textbackslash 9. Numerical simulations have been performed on the Swansea SUNBIRD system. This  system is part of the Supercomputing Wales project, which is part-funded by the European Regional Development Fund (ERDF) via Welsh Government.

\end{document}